\pdfoutput=1

\documentclass[11pt]{article}

\usepackage[]{acl}

\usepackage{enumitem}
\usepackage{times}
\usepackage{latexsym}

\usepackage[T1]{fontenc}

\usepackage[utf8]{inputenc}

\usepackage{microtype}

\usepackage{amsmath,amsfonts,bm}

\def\eqref#1{equation~\ref{#1}}

\def\1{\bm{1}}

\DeclareMathAlphabet{\mathsfit}{\encodingdefault}{\sfdefault}{m}{sl}
\SetMathAlphabet{\mathsfit}{bold}{\encodingdefault}{\sfdefault}{bx}{n}

\makeatletter

\newcommand*\totht[1]{\dimexpr\ht#1+\dp#1\relax}
\newcommand*\leading{{\setbox0\hbox{\strut}\the\totht0}}
\newcommand*\fntsize{{\setbox0\hbox{Mg}\the\totht0}}
\newcommand*\showsize[1]{{#1 {\ttfamily\string#1} (\f@size pt) \fntsize/\leading}\par}
\makeatother

\title{ZJUKLAB at SemEval-2025 Task 4: Unlearning via Model Merging}

\usepackage{microtype}
\usepackage{bbm}
\usepackage{amsmath}
\usepackage{amsfonts}
\usepackage{algorithm}
\usepackage{algpseudocode}
\usepackage{booktabs}
\usepackage{xparse}
\usepackage{graphicx}
\usepackage{multicol}
\usepackage{multirow}
\usepackage{tikz}
\usetikzlibrary{shapes,arrows}
\usepackage{natbib}
\usepackage{caption}
\usepackage{subcaption}
\usepackage{tikz-dependency}
\usepackage{wrapfig}
\usepackage{pgfplots}
\pgfplotsset{compat=1.17} 
\usepackage[T1]{fontenc}
\usepackage{subfiles}
\usepackage{readarray}
\usepackage{xcolor}
\usepackage{import}
\usepackage{hhline}
\usepackage{enumitem}
\usepackage{boldline}
\usepackage{mathtools}
\usepackage{colortbl}
\usepackage{cleveref}
\usepackage{fontawesome}
\usepackage[utf8]{inputenc}
\usepackage{graphicx}
\usepackage{tcolorbox}

\crefname{section}{§}{§§}
\Crefname{section}{§}{§§}

\usepackage{pifont}
\definecolor{bg}{rgb}{0.95, 0.95, 0.95}

\newcommand{\RNum}[1]{\uppercase\expandafter{\romannumeral #1\relax}}

\author{
    \textbf{Haoming Xu}\textsuperscript{1 \thanks{\quad Equal contribution}},  
    \textbf{Shuxun Wang}\textsuperscript{1 \footnotemark[1]}, 
    \textbf{Yanqiu Zhao}\textsuperscript{1 \footnotemark[1]}, \\
    \textbf{Yi Zhong}\textsuperscript{1 \footnotemark[1]}, 
    \textbf{Ziyan Jiang}\textsuperscript{1 \footnotemark[1]}, 
    \textbf{Ningyuan Zhao}\textsuperscript{1 \footnotemark[1]}, \\
    \textbf{Shumin Deng}\textsuperscript{2}, 
    \textbf{Huajun Chen}\textsuperscript{1}, 
    \textbf{Ningyu Zhang}\textsuperscript{1 \thanks{\quad Corresponding authors.}} \\
    \textsuperscript{1} Zhejiang University \quad \textsuperscript{2} National University of Singapore \\
    \texttt{haomingxu2003@gmail.com, zhangningyu@zju.edu.cn}
}

\begin{document}
\maketitle
\begin{abstract}
This paper presents the ZJUKLAB team's submission for \emph{SemEval-2025 Task 4: Unlearning Sensitive Content from Large Language Models}. This task aims to selectively erase sensitive knowledge from large language models, avoiding both over-forgetting and under-forgetting issues. We propose an unlearning system that leverages Model Merging (specifically TIES-Merging), combining two specialized models into a more balanced unlearned model. Our system achieves competitive results, ranking \textbf{second among 26 teams}, with an online score of 0.944 for Task Aggregate and 0.487 for overall Aggregate. In this paper, we also conduct local experiments and perform a comprehensive analysis of the unlearning process, examining performance trajectories, loss dynamics, and weight perspectives, along with several supplementary experiments, to understand the effectiveness of our method. Furthermore, we analyze the shortcomings of our method and evaluation metrics, emphasizing that MIA scores and ROUGE-based metrics alone are insufficient to fully evaluate successful unlearning. Finally, we emphasize the need for more comprehensive evaluation methodologies and rethinking of unlearning objectives in future research\footnote{\quad Code is available at \url{https://github.com/zjunlp/unlearn/tree/main/semeval25}.}.
\end{abstract}

\section{Introduction}
\label{intro}
Unlearning has emerged as a critical technique in AI systems, enabling the selective removal of sensitive data, including copyrighted material and personal information, from trained models.
As the International AI Safety Report \citep{bengio2025internationalaisafetyreport} emphasizes, unlearning plays a vital role in mitigating privacy and copyright risks associated with extensive training datasets.
However, it also acknowledges that current unlearning methods remain inadequate, which often fail to completely erase targeted data while potentially degrading model performance, thus limiting practical implementation.

Specifically, existing unlearning methods often struggle with over-forgetting (excessive elimination of non-sensitive information) or under-forgetting (incomplete removal of sensitive data).
It is challenging to find optimal hyperparameters that balance performance across multiple evaluation dimensions, sometimes even impossible.
To address these limitations, we propose a novel unlearning system that leverages model merging to combine an over-forgetting model with an under-forgetting model, creating a more effective unlearned model. 
It can produce superior results simply by merging two models with complementary biases.

Our system achieved second place in \emph{SemEval-2025 Task 4: Unlearning Sensitive Content from Large Language Models}, with our 7B model attaining a \emph{Task Aggregate Score} of 0.944 and \emph{Aggregate Score} of 0.487,  demonstrating the effectiveness of our system in selectively removing sensitive content.
Furthermore, our local experiments yielded almost perfect results with a \emph{MIA Score} of 0.501 and \emph{Aggregate Score} of 0.806, while maintaining an exceptionally high \emph{Task Aggregate} and comparable \emph{MMLU Avg.}. 
We provide comprehensive analyses that validate our system's effectiveness and offer deeper insights into the unlearning process.

\begin{figure*}[!htbp]
\includegraphics[width=\linewidth]{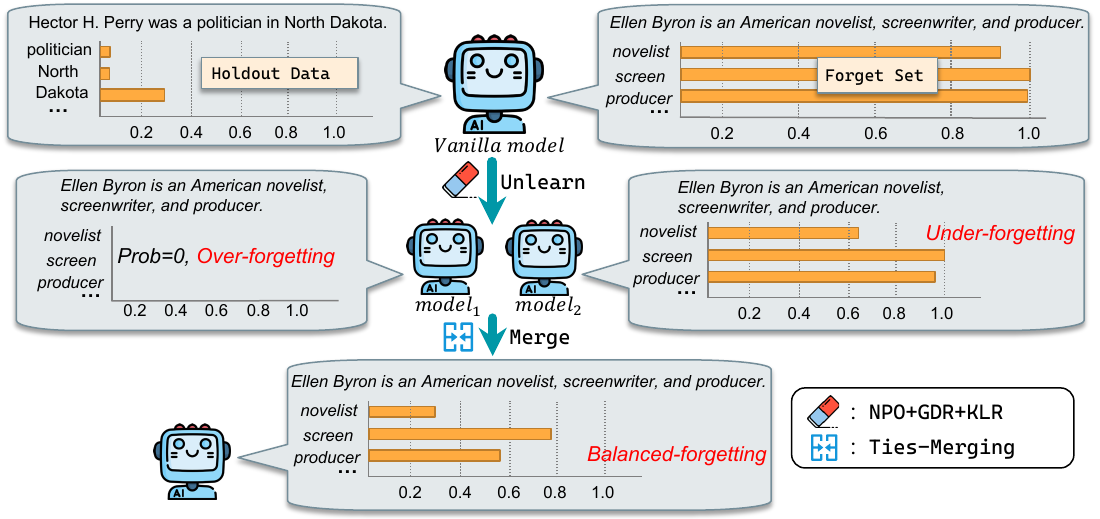}
\caption{Visualizing Unlearning via Model Merging. 
The vanilla model (top) initially assigns high probabilities to forget set (member) and low probabilities to holdout data (nonmember). 
We then merge two individually unlearned models: 
one exhibiting over-forgetting (middle left) and the other under-forgetting (middle right). 
Model merging aims to achieve balanced forgetting (bottom), effectively reducing the model's confidence in predicting sensitive member data while preserving its performance on nonmember data.}
\label{fig:methods}
\end{figure*}
\section{Task Description}
\noindent\textbf{Datasets}\quad
The dataset comprises a \emph{forget set} and a \emph{retain set} across three subtasks: 
(1) long-form synthetic creative documents, 
(2) short-form synthetic biographies with PII (names, phone numbers, SSN, emails, addresses), 
and (3) real documents from the target model's training data. 
The organizers provide a vanilla model (OLMo-7B-0724-Instruct)~\citep{Groeneveld2023OLMo} which has been pretrained on all subtasks.

\noindent\textbf{Evaluation}\quad
\label{sec:evaluation}
Evaluation involves sentence completion and question answering across tasks. 
Key metrics include: 
\emph{Regurgitation Score} (ROUGE-L for sentence completion), \emph{Knowledge Score} (accuracy for QA), 
\emph{MIA Score} (loss-based membership inference attack \citep{shi2024detectingpretrainingdatalarge}), and \emph{MMLU Score} (average accuracy on 57 STEM subjects). 
\emph{Task Aggregate} is the harmonic mean of Regurgitation Scores and Knowledge Scores for each task. 
The overall \emph{Aggregate} averages the Task Aggregate, MIA scores, and MMLU scores.  

For details about task description, please refer to the official paper \citep{ramakrishna2025lumellmunlearningmultitask,ramakrishna2025semeval2025task4unlearning}.










\section{Methodology} 
\label{sec:methods}
As illustrated in Figure \ref{fig:methods}, our unlearning system follows two phases. 
(1) the \emph{Training Phase} develops two complementary models, each exhibiting strong performance. 
(2) the \emph{Merging Phase} merges these models, leveraging their strengths to achieve effective and balanced unlearning.
\subsection{Training Phase}
We train two models with identical objectives but different hyperparameters via Low-Rank Adaptation (LoRA) \citep{hu2021loralowrankadaptationlarge}. 
Three components are included in the optimization process: \textbf{Negative Preference Optimization (NPO)} \citep{npo} on forget set, alongside \textbf{Gradient Descent on Retain Set (GDR)} and \textbf{Kullback-Leibler Divergence Minimization on Retain Set (KLR)}.
The composite objective is as follows:
\vspace{-1ex}
\begin{equation}
L_{\text{total}} = \alpha L_{\text{npo}} + \beta L_{\text{gdr}} + \gamma L_{\text{klr}},
\end{equation}
where $L_{\text{npo}}$ leverages the preference optimization to minimize probabilities of target tokens on forget data, while $L_{\text{gdr}}$ and $L_{\text{klr}}$ preserve retain data. 
The hyperparameters \(\alpha, \beta, \gamma\) are set to balance forgetting and retention.
Our aim is to train two complementary models that exhibit distinct strengths in metrics.
Detailed formulations are shown in Appendix~\ref{appendix:baselines}.
\subsection{Merging Phase}
After training, we apply \textbf{TIES-Merging} \citep{yadav2023tiesmergingresolvinginterferencemerging} to combine the LoRA adapters of the two models. This involves three stages:

\textbf{Trimming}: Preserving only the most significant parameters based on a density threshold while zeroing out the rest.

\textbf{Electing}: Creating a unified sign vector that resolves parameter conflicts by identifying the dominant direction of change across models.

\textbf{Disjoint Merging}: Averaging non-zero parameter values that align with the unified sign vector, ensuring that the merged model incorporates only changes contributing to the agreed direction, thus improving multitask performance.
\begin{table*}[ht]
    \centering
    \resizebox{\textwidth}{!}{
    \begin{tabular}{l|c|c|ccc}
        \toprule
        Environment & Algorithm & Aggregate & Task Aggregate & MIA Score/MIA AUC & MMLU Avg. \\
        \midrule
        \multirow{2}{*}{Online} 
        & AILS-NTUA & \textbf{0.706} & 0.827 & \textbf{0.847} / -- & 0.443 \\
        & YNU & 0.470 & 0.834 & 0.139 / -- & 0.436 \\
        & Mr.Snuffleupagus & 0.376 & 0.387 & 0.256 / -- & \textbf{0.485} \\
        & ZJUKLAB (ours) & 0.487 & \textbf{0.944} & 0.048 / -- & 0.471\\
        \midrule\midrule
        \multirow{3}{*}{Local} 
        & NPO+GDR+KLR ($model_1$) & 0.481 & 0.968 & 0.045 / $0.022^{\clubsuit}$ & 0.431 \\
        & NPO+GDR+KLR ($model_2$) & 0.504 & 0.659 & 0.364 / $0.818^{\spadesuit}$ & 0.491 \\
        & Ours & \textbf{0.806} & \textbf{0.939} & \textbf{0.997} / $0.501^{\heartsuit}$ & \textbf{0.480} \\
        \bottomrule
    \end{tabular}
    }
    \caption{The online and local experiments results. Note that $\clubsuit$ indicates over-forgetting, $\spadesuit$ indicates under-forgetting, and $\heartsuit$ signifies balanced forgetting, achieving a raw MIA AUC close to 0.5. All metrics are detailed in \S\ref{sec:evaluation}.}
    \label{tb:baselines}
\end{table*}

\section{Experiments} 
\label{sec:exp}
\paragraph{Implementation}
We carried out our experiments using two NVIDIA A100-PCIE-40GB GPUs.
The organizers supplied the local dataset for our local experiments and evaluated our code online using an additional unreleased dataset. 
Detailed configurations are provided in Appendix \ref{sec:config}.

\paragraph{Main Results}
Table \ref{tb:baselines} presents the online results evaluated by the organizers and the local results evaluated by us.
Our 7B model achieves an \emph{Aggregate} score of 0.487 online, ranking second among 26 teams. 
The online \emph{MIA Score} is less favorable, possibly due to dataset discrepancies between the online and local environments. 
However, local evaluations effectively validate the core principles of our system design. 
In \emph{training phase}, $model_1$ shows over-forgetting, achieving a high \emph{Task Aggregate} of 0.968 but a low \emph{MIA Score} of 0.022.  
In contrast, $model_2$ shows under-forgetting, with a lower \emph{Task Aggregate} of 0.659 and a higher \emph{MIA Score} of 0.818. 
The merged model shows better performance, attaining a \emph{Task Aggregate} of 0.939 and a \emph{MIA AUC} of 0.501.
This merging technique integrates the strengths of both models, preserving their high \emph{Task Aggregate} and \emph{MMLU Avg.} scores while successfully neutralizing their MIA scores, resulting in an almost ideal MIA score. 
These results highlight our system's ability to effectively aggregate the strengths of these biased models. 

\section{Analysis}
\subsection{Why NPO+GDR+KLR Works?}
This section analyzes the effectiveness of NPO+GDR+KLR model (denoted as $model_1$ in the training phase), trained on the local dataset.

\paragraph{Performance Trajectory}
\begin{figure}[!t]
\includegraphics[width=\linewidth]{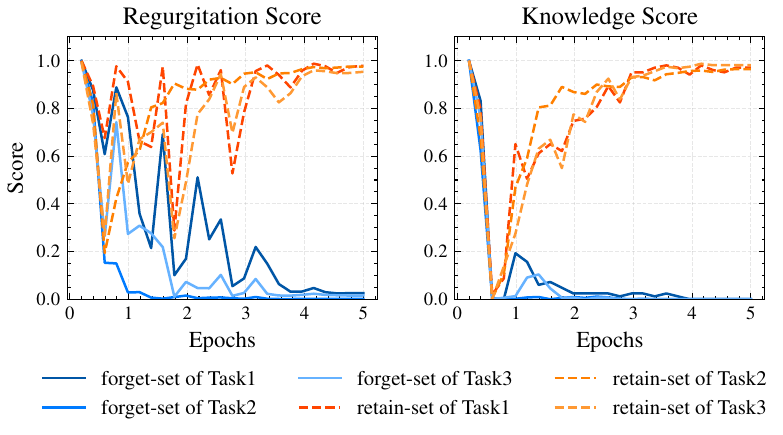}
\caption{Performance Curves: Regurgitation and Knowledge Scores During Training.}
\label{fig:performance}
\end{figure}
\begin{figure}[t]
\includegraphics[width=\linewidth]{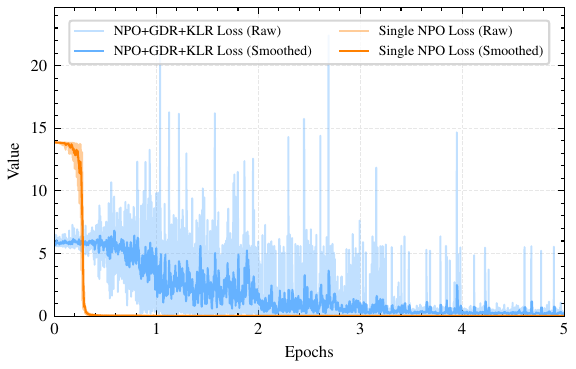}
\caption{Training Loss Curves of NPO and NPO+GDR+KLR.}
\label{fig:loss}
\end{figure}
To understand performance trends, we evaluated model checkpoints throughout training.
As shown in Figure \ref{fig:performance}, both Regurgitation and Knowledge Scores initially decline concurrently for forget and retain sets (epochs 0-0.8). 
This suggests that, in the early stages of training, the optimization processes for both forgetting and retaining knowledge are proceeding in the same direction, causing a simultaneous metric decrease.
Subsequently, the Knowledge Score steadily trends upward, while the Regurgitation Score increases with noticeable oscillations.
This indicates that the optimization directions of knowledge retention and knowledge forgetting are beginning to become different.
The observed fluctuations in Regurgitation Score may stem from the tradeoff between learning and forgetting.

\paragraph{Loss Dynamics}
Figure \ref{fig:loss} compares the training loss curves of NPO and NPO+GDR+KLR models.
Notably, the NPO+GDR+KLR loss curve displays oscillations in mid-training, likely caused by the similarity between forget and retain sets, hindering a steady loss decline.
Conversely, training with only the NPO loss function results in rapid convergence and a smooth loss curve, further highlighting the conflict between NPO and regularization. 
Despite this, NPO+GDR+KLR achieves a stable loss value in later training stages, demonstrating its ability to effectively balance forgetting and retention.
\begin{figure}[!t]
\includegraphics[width=\linewidth]{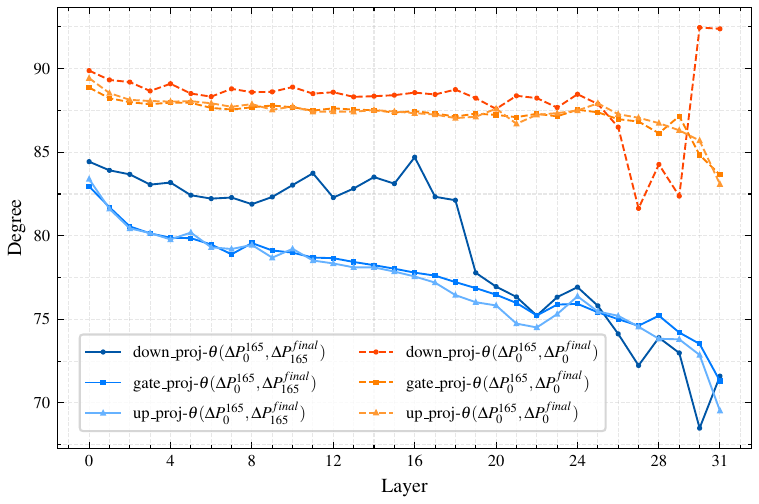}
\caption{Angle ($\theta$) between Parameter Change Vectors: $\Delta P_{0}^{165}$, $\Delta P_{165}^{final}$, $\Delta P_{0}^{final}$. }
\label{fig:weight}
\end{figure}
\paragraph{Weight perspective}
Figure \ref{fig:performance} shows a performance trend with an initial decline followed by an increase. 
We identify this turning point as the \textit{inflection point} (step 165). 
To understand optimization dynamics around this point, we analyzed the angle between flattened parameter change vectors across training phases (Figure~\ref{fig:weight}), where $\Delta P_{s_1}^{s_2}$ be the parameter change vector from step $s_1$ to $s_2$.
The angle between $\Delta P_{0}^{165}$ and $\Delta P_{165}^{final}$ is approximately 70-85 degrees.
This suggests that the initial phase overemphasizes forgetting, while a significant shift in optimization direction occurs after the inflection point, where the balance between forgetting and retention has gradually been established. 
Conversely, the angle between the initial direction ($\Delta P_{0}^{165}$) and the overall optimization direction ($\Delta P_{0}^{final}$) approaches near orthogonality (90 degrees).
This indicates that overall training does not consistently follow the initial direction, and the initial "forgetting" emphasis is balanced by later retention optimization. 

\subsection{Why Merge works?}
\begin{table}[!t]
\centering
\begin{tabular}{c|c}
    \toprule
    Merging methods & Agggregate \\
    \midrule
    Linear & 0.244 \\
    DARE-Linear & 0.440 \\
    DARE-TIES & 0.561 \\
    Magnitude Prune & 0.558 \\
    \textbf{TIES} & \textbf{0.806} \\
    \bottomrule
\end{tabular}
\caption{Merging techniques comparison}
\label{tb:merge}
\end{table}
\begin{figure}[!t]
\includegraphics[width=\linewidth]{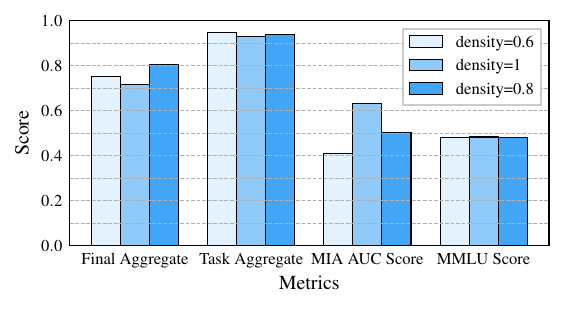}
\caption{Performance for different density choices}
\label{fig:desity}
\end{figure}
To understand the efficacy of merging, we conduct comparative experiments on different merging techniques.
As shown in Table~\ref{tb:merge}, TIES-Merging outperforms others, and this effectiveness comes from its three fundamental operations:  \textbf{Trim}, \textbf{Elect}, and \textbf{Disjoint Merge}. 

Firstly, for \textbf{Trimming}, we conduct ablation studies varying the density between 0.6, 0.8, and 1 (Figure~\ref{fig:desity}) and observed that a density of 0.8 yields the best results. 
This optimal density level retains essential parameters while removing redundant ones, effectively preserving the better performance of two models and achieving balanced forgetting.
We hypothesize that lower densities (e.g., 0.6) excessively prune parameters vital for knowledge retention, leading to over-unlearning and a reduced MIA score. 
Conversely, a density of 1, by retaining all parameters, introduces redundancy and may incorporate influences from the less-unlearned model, resulting in a suboptimal outcome and a higher MIA score. 
Therefore, trimming with a density of 0.8 strikes a critical balance.
Beyond trimming, TIES-Merging further enhances directional consistency through the \textbf{Elect} operation, which establishes parameter signs based on magnitude. Given the strong baseline performance of the individual models, this magnitude-based election ensures reliable convergence toward optimal directional consistency during merging. 
Finally, the \textbf{Disjoint Merging} operation averages parameters with consistent elected signs and discards discordant ones. This strategic approach effectively mitigates over-unlearning and further enhances the merged model's resistance to Membership Inference Attacks (MIA).

\section{Rethinking Unlearning}
\subsection{Drawbacks: Over-forgetting Phenomena}
\label{Drawbacks}
Despite demonstrating effectiveness, our system still exhibits \textbf{over-forgetting}.
Firstly, the unlearned model exhibits \emph{model collapse}, frequently generating repetitive characters (e.g., "6 6 6").
This phenomenon arises from the training process itself, the model may find a suboptimal but easy shortcut: generating repetitive outputs to reduce loss.
Specifically, Task 2 involves a digit-heavy dataset, so the model will take this high-frequency option as their outputs.
Secondly, we observed \textbf{forgetting of generic knowledge}.  
We analyze question patterns of forget set to construct 50 common knowledge questions (e.g., "What is the capital of France?"), finding a significant Knowledge Score drop (0.88 $\to$ 0.35) against a vanilla baseline.
These drawbacks are also observed in some studies \citet{mekala-etal-2025-alternate, xu2025relearnunlearninglearninglarge}, highlighting a fundamental weakness of this paradigm.  
Throughout training, the system repeatedly applies reverse optimization signals to the original forget data. Without positive guidance like in reinforcement learning, the model cannot explore better outputs and inevitably degrades under sustained pressure.
\begin{tcolorbox}[
    colback=gray!5,
    colframe=gray!40,
    fonttitle=\bfseries\footnotesize,
    top=0.5mm,        
    bottom=0.5mm,     
    left=0.5mm,       
    right=0.5mm,      
    boxsep=0.5mm,     
    arc=0.5mm,        
    boxrule=0.5mm,  
    ]
\footnotesize
\setlength{\parskip}{0pt}
\setlength{\parsep}{0pt}

\begin{minipage}{\linewidth}
\begin{tcolorbox}[
    colback=white,
    colframe=red!70!black,
    leftrule=0.5mm,
    rightrule=0mm,
    toprule=0mm,
    bottomrule=0mm,
    arc=0mm,
    top=0.5mm,
    bottom=0.5mm,
    left=1mm,
    right=0.5mm,
    boxsep=0mm,
]
\textbf{\textcolor{red!70!black}{Forget Set Case:}} \\
\texttt{\textbf{Question}: What is Lorette Fuchsia's email address?} \\
\texttt{\textbf{Answer}:  6 6 6 6 6 6 6...}
\end{tcolorbox}
\end{minipage}

\vspace{0.5mm}

\begin{minipage}{\linewidth}
\begin{tcolorbox}[
    colback=white,
    colframe=green!60!black,
    leftrule=0.5mm,
    rightrule=0mm,
    toprule=0mm,
    bottomrule=0mm,
    arc=0mm,
    top=0.5mm,
    bottom=0.5mm,
    left=1mm,
    right=0.5mm,
    boxsep=0mm,
]
\textbf{\textcolor{green!60!black}{Retain Set Case:}} \\
\texttt{\textbf{Question}: What is the birth date of Fredericka Amber?} \\
\texttt{\textbf{Answer}:  1969-12-21}
\end{tcolorbox}
\end{minipage}

\vspace{0.5mm}

\begin{minipage}{\linewidth}
\begin{tcolorbox}[
    colback=white,
    colframe=blue!70!black,
    leftrule=0.5mm,
    rightrule=0mm,
    toprule=0mm,
    bottomrule=0mm,
    arc=0mm,
    top=0.5mm,
    bottom=0.5mm,
    left=1mm,
    right=0.5mm,
    boxsep=0mm,
]
\textbf{\textcolor{blue!70!black}{Generic Knowledge Case:}} \\
\texttt{\textbf{Question}: In which city is the Eiffel Tower located?} \\
\texttt{\textbf{Answer}:  6 6 6 6 6 6 6...}
\end{tcolorbox}
\end{minipage}
\end{tcolorbox}

\subsection{Limitations of Unlearning Evaluation}
\textbf{ROUGE-based metrics} primarily measure how closely a response matches an expected output rather than exact knowledge unlearning. 
For instance, a different long response might still inadvertently leak sensitive information like an email address, yet escape detection by ROUGE-L due to its focus on textual overlap rather than content semantics.
In this competition, separate metrics have been introduced (i.e., Regurgitation Score and Knowledge Score). 
However, they remain susceptible to superficial textual variations, where minor rephrasing can mask underlying retention of knowledge, thus undermining their ability to accurately evaluate unlearning effectiveness.
Similarly, \textbf{MIA Scores like Min-k\% prove insufficient}.  
Although our method achieves an almost optimal MIA score of 0.501, it still generates repetitive outputs that deviate from the base model's behavior.  
Some studies \citep{duan2024membershipinferenceattackswork, meeus2024sokmembershipinferenceattacks} cast doubt on MIA's reliability for LLMs, pointing to potential temporal or domain discrepancies in datasets.  
In this competition, while the forget set and retain set are derived from Wikipedia after the deadline of OLMO's training, subtle distribution shifts may still persist.  
Our local test on OLMo-7B-0724-Instruct-hf yields an MIA AUC of 0.46, slightly misaligned with the official optimal score of 0.5, further highlighting these inconsistencies.

\subsection{Rethinking Unlearning's Objectives}
Recent studies \citep{10488864, zhou2024limitationsprospectsmachineunlearning, thaker2024positionllmunlearningbenchmarks, cooper2024machineunlearningdoesntthink, barez2025openproblemsmachineunlearning} present critical analyses of generative AI unlearning. 
These studies collectively reveal three fundamental limitations: 
(1) current unlearning methods remain impractical, 
(2) evaluations fail to assess the generalization capability of unlearned models, 
and (3) benchmarks encourage model to overfit the training set, creating an illusory forgetting.
The root challenge lies in the lack of a clearly defined, universally applicable unlearning objective. Rather than overloading unlearning with goals like resistance to relearning attacks \citep{fan2025llmunlearningresilientrelearning}, future research should prioritize on-demand unlearning and robust evaluation to address practical policy needs.
As discussed in \S\ref{Drawbacks}, current methods often lead to degraded outputs. 
Future work can explore the incorporation of positive signals to guide the model toward more appropriate forgetting behaviors such as data augmentation and reinforcement learning.

\section{Conclusion}
This paper introduce an unlearning system via model merging. 
By combining two complementary models, it effectively achieves balanced forgetting and excellent knowledge preservation.

\section*{Acknowledgements}

We would like to express our great gratitude to the anonymous reviewers for their kind comments.
This work was supported by the National Natural Science Foundation of China (No. 62206246), the Fundamental Research Funds for the Central Universities (226-2023-00138), Yongjiang Talent Introduction Programme (2021A-156-G), CIPSC-SMP-Zhipu Large Model Cross-Disciplinary Fund, Tencent AI Lab Rhino-Bird Focused Research Program (RBFR2024003).
We gratefully acknowledge the support of Zhejiang University Education Foundation Qizhen Scholar Foundation.

\bibliography{anthology,main}

\appendix

\section{Detailed Setup}
\subsection{Detailed formulas}
\label{appendix:baselines}
This section introduces detailed formulas in this paper.

\textit{Negative Preference Optimization}: The loss function penalizes the model for generating outputs with negative preferences while maximizing outputs with positive preferences:
\begin{equation}
    L_{\text{NPO}} = -\frac{2}{\beta} \mathbb{E}_{\mathcal{D}_f} \left[ \log \sigma \left( -\beta \log \frac{\pi_\theta(y|x)}{\pi_{\text{ref}}(y|x)} \right) \right]
\end{equation}
where \( \pi_\theta(y|x) \) is the model's output distribution and \( \pi_{\text{ref}}(y|x) \) is the reference distribution. Here, \( \sigma \) is the sigmoid function and \( \beta \) is a regularization parameter.

\textit{Gradient Descent on Retain Set (GDR)}: Minimizes the loss for samples in retain set by updating parameters in the direction to the gradient of the loss function:
\begin{equation}
    \theta_{t+1} = \theta_t - \eta \nabla_\theta \mathcal{L}(\theta_t, \mathcal{D}_{r})
\end{equation}
where \( \theta_t \) represents the model parameters at step \( t \), \( \eta \) is the learning rate, and \( \nabla_\theta \mathcal{L}(\theta_t, \mathcal{D}_{r}) \) is the gradient of the loss function at step \( t \), calculated on the retain set \( \mathcal{D}_{r} \).

\textit{KL Minimization on Retain Set (KLR)}: Minimizes the Kullback-Leibler divergence between the model’s output distribution and a target distribution on retain set:
\begin{equation}
    \mathcal{L}_{\text{klr}} = \sum_{i} \pi_{\theta}(y_i) \log \frac{\pi_{\theta}(y_i)}{\pi_{\text{target}}(y_i | \mathcal{D}_{r})}
\end{equation}
where \( \mathcal{L}_{\text{KL}} \) is the loss, \( \pi_{\theta}(y_i) \) is the model's output distribution for the \( i \)-th output token \( y_i \), and \( \pi_{\text{target}}(y_i | \mathcal{D}_{r}) \) is the target output distribution for the \( i \)-th token \( y_i \) conditioned on the retain set \( \mathcal{D}_{r} \).

    

\subsection{Detailed Implementation}
\label{sec:config}
Table~\ref{tab:config} summarizes the complete configuration parameters used in our experiments. 

\begin{table}[htbp]
\centering
\begin{tabular}{@{}lccc@{}}
\toprule
\textbf{Parameter} & \textbf{Model$_1$} & \textbf{Model$_2$} \\ \midrule
batch\_size  & 1 & 2  \\
gradient\_accumulation & 4 & 4 \\
num\_epochs & 5 & 5 \\
lr & $1\times10^{-4}$ & $1\times10^{-4}$   \\
max\_length & 256 & 256 \\
weight\_decay & 0.01 & 0.01 \\
seed & 42  & 42 \\
ga\_ratio  & 0.4 & 0.3 \\
gd\_ratio & 0.4 & 0.3 \\
gk\_ratio & 0.2 & 0.4 \\
LoRA\_r & 32 & 32 \\
LoRA\_alpha  & 32 & 32 \\
LoRA\_dropout & 0.05  & 0.05 \\
\bottomrule
\end{tabular}
\caption{Complete Hyperparameters Configuration.}
\label{tab:config}
\end{table}

\section{Related Work}
\paragraph{LLM Unlearning}
The topic of unlearning in large language models \citep{Chen_2024} has recently attracted significant attention in the literature.
One approach to unlearning is Gradient Ascent \citep{ga}, which aims to maximize the loss function to facilitate forgetting. Another method, Negative Preference Optimization (NPO) \citep{npo}, builds upon Direct Preference Optimization (DPO) \citep{DPO}, offering an alternative strategy for model unlearning.
Various unlearning techniques have been proposed, including those presented by \citep{NEURIPS2022_b125999b, eldan2023whosharrypotterapproximate, yu-etal-2023-unlearning, chen2023unlearnwantforgetefficient, 10.1007/978-981-97-9443-0_22, gandikota2024erasingconceptualknowledgelanguage, jiang2025anyediteditknowledgeencoded, liu-etal-2024-towards-safer, zhuang2024uoeunlearningexpertmixtureofexperts}.
An alternative strategy, referred to as “locate-then-unlearn,” is exemplified by KnowUnDo \citep{tian2024forgetnotpracticalknowledge} and SURE \citep{zhang2024doesllmtrulyunlearn}, which focus on knowledge localization before executing the unlearning process.
Additionally, data-driven methods for unlearning have also been introduced, such as those proposed by \citep{jang2022knowledgeunlearningmitigatingprivacy, ma2024unveilingentitylevelunlearninglarge, liu2024learningrefusemitigatingprivacy, gu2024meowmemorysupervisedllm, sinha2024unstarunlearningselftaughtantisample, xu2025relearnunlearninglearninglarge, mekala-etal-2025-alternate}.
Several works have explored the use of model merging techniques to achieve unlearning \citep{kadhe2024splitunlearnmergeleveraging, kuo2025exactunlearningfinetuningdata}.

\paragraph{Model Merging}
Training a model for each task can be costly, but model merging offers a solution to these challenges by combining multiple pre-trained models.
Model merging strategies include parameter averaging (Linear), singular value decomposition (SVD) for low-rank alignment, and feature concatenation (CAT). 
Advanced variants like TIES \citep{yadav2023tiesmergingresolvinginterferencemerging} trim redundant parameters and resolve sign conflicts, while TIES-SVD \citep{stoica2024modelmergingsvdtie} integrates SVD for refined fusion. 
DARE methods\citep{yu2024languagemodelssupermario}, and methods like DARE-TIES, DARE-linear introduce parameter dropout and rescaling, with extensions (DARE-TIES-SVD, DARE-linear-SVD) combining SVD for structured compression. 
The magnitude-prune \citep{deep2024dellamergingreducinginterferencemodel} removes low-impact weights, and its SVD variant (magnitude-prune-SVD) is further compressed via low-rank decomposition.

\end{document}